\documentclass[10pt,twocolumn,letterpaper]{article}
\usepackage{cvpr}
\usepackage{times}
\usepackage{epsfig}
\usepackage{subfigure}
\usepackage[table]{xcolor}
\usepackage[noblocks]{authblk}
\usepackage{graphicx}
\usepackage{amsmath}
\usepackage{amssymb}

\usepackage[pagebackref=true,breaklinks=true,letterpaper=true,colorlinks,bookmarks=false]{hyperref}

\cvprfinalcopy 

\def\cvprPaperID{****} % *** Enter the CVPR Paper ID here

\ifcvprfinal\fi
\begin{document}

%%%%%%%%% TITLE
\title{3D Correspondence Grouping with Compatibility Features}

\author[1]{Jiaqi Yang}
\author[1]{Jiahao Chen}
\author[1]{Zhiqiang Huang}
\author[2]{Siwen Quan}
\author[1]{Yanning Zhang}
\author[3]{Zhiguo Cao\thanks{Corresponding author}}
\affil[1]{\small School of Computer Science, Northwestern Polytechnical University}
\affil[2]{ School of Electronic and Control Engineering, Chang'an University}
\affil[3]{ School of Artificial Intelligence and Automation, Huazhong University of Science and Technology}
\maketitle
\pagestyle{empty}
\thispagestyle{empty}

%%%%%%%%% ABSTRACT
\begin{abstract}
We present a simple yet effective method for 3D correspondence grouping. The objective is to accurately classify initial correspondences obtained by matching local geometric descriptors into inliers and outliers. Although the spatial distribution of correspondences is irregular, inliers are expected to be geometrically compatible with each other. Based on such observation, we propose a novel representation for 3D correspondences, dubbed compatibility feature (CF), to describe the consistencies within inliers and inconsistencies within outliers. CF consists of top-ranked compatibility scores of a candidate to other correspondences, which purely relies on robust and rotation-invariant geometric constraints. We then formulate the grouping problem as a classification problem for CF features, which is accomplished via a simple multilayer perceptron (MLP) network. Comparisons with nine state-of-the-art methods on four benchmarks demonstrate that: 1) CF is distinctive, robust, and rotation-invariant; 2) our CF-based method achieves the best overall performance and holds good generalization ability.
\end{abstract}

%%%%%%%%% BODY TEXT
\section{Introduction}
3D correspondence grouping (a.k.a. 3D correspondence selection or 3D mismatch removal) is essential to a number of point-to-point correspondences-based tasks, such as 3D point cloud registration~\cite{rusu2009fast}, 3D object recognition~\cite{tombari2010unique}, and 3D reconstruction~\cite{mian2005automatic}. The aim is to classify initial feature correspondences between two 3D point clouds obtained by matching local geometric descriptors into inliers and outliers. Due to a number of factors, e.g., repetitive patterns, keypoint localization errors, and data nuisances including noise, limited overlap, clutter and occlusion, heavy outliers are generated in the initial correspondence set~\cite{Yang2020corr_group_eval}. Thus, it is very challenging to mine the consistency of scarce inliers and find those inliers.
\begin{figure}[t]
	\centering
	\includegraphics[width=\linewidth]{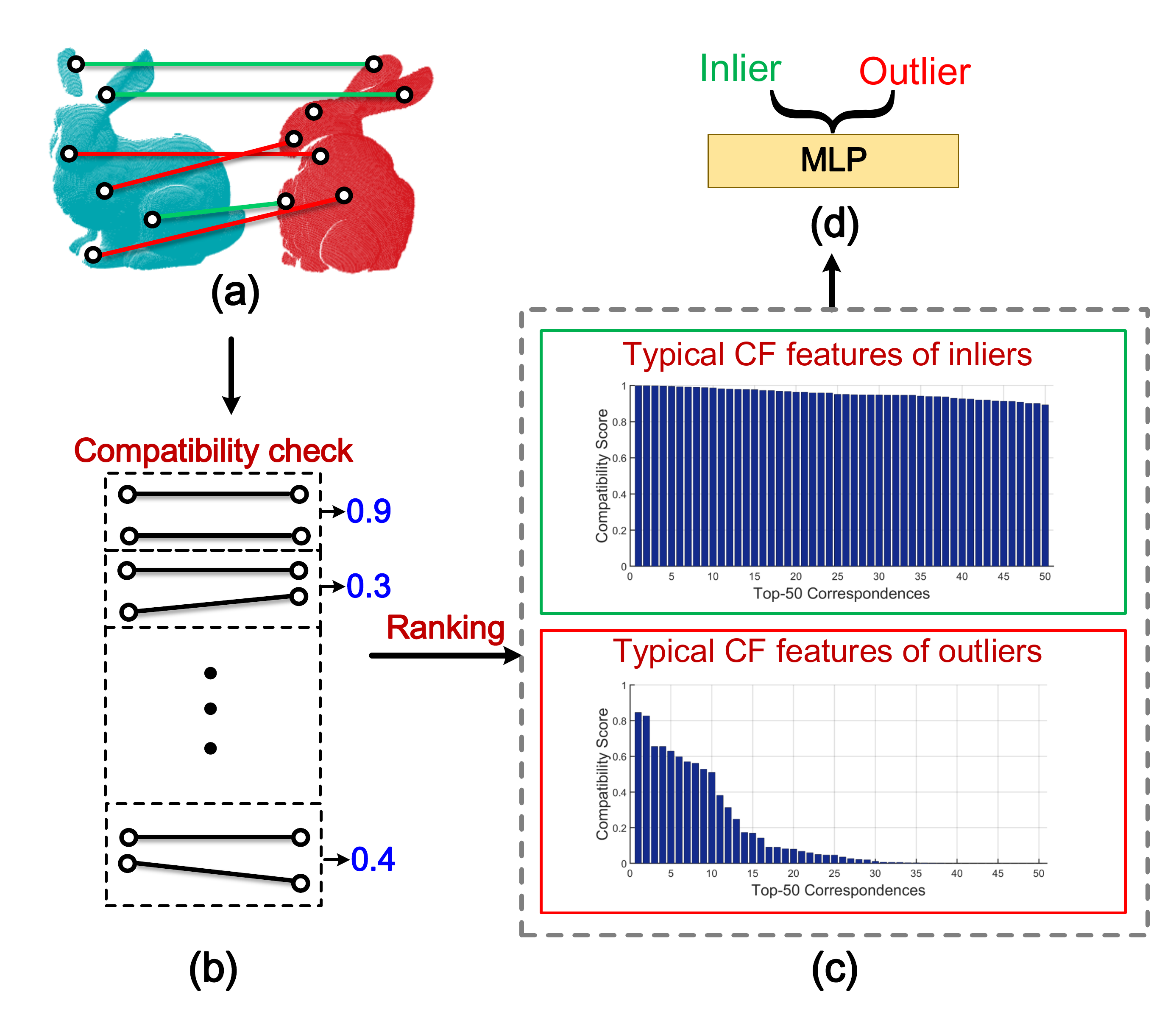}\\
	\caption{Illustration of the proposed CF-based method for 3D correspondence grouping. {\bf(a)} 3D point-to-point feature correspondences between two point clouds. {\bf(b)} The geometrical compatibility scores of each correspondence with others are computed. {\bf(c)} Typical CF features of inliers and outliers, which indicate the discriminative power of CF. {\bf(d)} CF features are fed to an MLP network for binary classification.}
	\label{fig:illus}
\end{figure}

Existing 3D correspondence grouping methods can be divided into two categories: group-based and individual-based. Group-based methods~\cite{fischler1981random,leordeanu2005spectral,chen20073d,tombari2010object,rodola2013scale} assume that inliers constitute a cluster in a particular domain and struggle to recover such cluster. By contrast, individual-based ones~\cite{mian2006three,yang2016fast,lowe2004distinctive,buch2014search,yang2019ranking,sahloul2020accurate} usually first assign confidence scores to correspondences based on feature or geometrics constraints, and then select top-scored correspondences independently. However, as revealed by a recent evaluation~\cite{Yang2020corr_group_eval}, existing methods in both categories {\textbf{1)}} generalize poorly across datasets with different application scenarios and data modalities, and {\textbf{2)}} deliver limited precision performance which is critical to successful 3D registration with sparse correspondences.

To overcome above limitations, we present a new feature presentation to describe 3D correspondences dubbed compatibility feature (CF) along with a CF-based 3D correspondence grouping method, as illustrated in Fig.~\ref{fig:illus}. CF consists of top-ranked compatibility scores of a candidate to other correspondences. CF is supposed to hold strong discriminative power because {\textit{inliers are geometrically compatible with each other whereas outliers are unlikely to be compatible with either outliers or inliers}} due to their unordered spatial distributions. This results in clear distinctions between CF features of inliers and outliers. Since the correspondence grouping problem can be viewed as a {\textit{binary classification problem}}, we train a simple multilayer perceptron (MLP) network as a robust classifier to distinguish inliers and outliers. Although there have been some ``end-to-end'' learning-based 2D correspondence selection methods~\cite{moo2018learning,zhang2019learning,sarlin2019superglue}, our method follows a ``geometry + learning'' fashion due to the following reasons. {\textbf{First,}} even for 2D images with pixel coordinate values being in a small range, training ``end-to-end'' networks still requires a huge amount number of labeled image pairs~\cite{moo2018learning}. By contrast, the coordinates of 3D points can be arbitrary in a 3D space, greatly increasing the challenges of  training data preparation and training. We will show that dozens of point cloud pairs are suffice to train an MLP to classify CF features. {\textbf{Second,}} pixel/point coordinates are sensitive to rotations~\cite{deng2018ppf}. Although augmenting training data can sometimes alleviates this problem, the network is still not fully rotation-invariant in nature. By contrast, CF features are extracted with rotation-invariant geometric constraints and are robust to arbitrary 3D rotations. {\textbf{Third,}} most of existing ``end-to-end'' methods are not practical on real-world data as demonstrated in~\cite{choy2020deep}. {\textbf{Fourth,}} with CF features, the learning network (i.e., MLP) in our method is very lightweight and can be trained with a few number of point cloud pairs. In a nutshell, this paper has the following contributions.
\begin{itemize}
	\item A compatibility feature (CF) representation is proposed to describe 3D feature correspondences. CF captures the key differences between inliers and outliers regarding pairwise geometrical compatibility, which is distinctive, robust, and rotation-invariant. 
	\item A 3D correspondence grouping method based on CF is proposed. In 3D correspondence grouping domain, our method is the first learning-based one (to the best of our knowledge), while it holds the ``geometry + learning'' property and works with a simple MLP network. Comprehensive experiments and comparisons with all methods evaluated in~\cite{Yang2020corr_group_eval} on datasets  with different application contexts and data modalities verify that our method has good generalization abilities and achieves outstanding precision performance. 
\end{itemize}

\section{Related Work}
This section briefly reviews group-based and individual-based methods for 3D correspondence grouping. Methods in both categories are geometric-only ones. Because our method includes a learning-based classier, we also discuss some learning-based techniques for correspondence problems in 2D domain. 
\subsection{3D Correspondence Grouping}
\noindent\textbf{Group-based methods}
Random sampling consensus~\cite{fischler1981random} is arguably the most commonly used method for 3D correspondence grouping and transformation estimation. It iteratively estimates a model from correspondences and verifies its rationality; correspondences  coherent with the best estimated model are served as inliers. The variants of RANSAC~\cite{guo2013rotational,quan2020_cgsac} generally follow the same pipeline. Some methods try to find the main cluster within initial correspondences by analyzing the affinity matrix computed for correspondences. For instance, game theory matching (GTM)~\cite{rodola2013scale} and spectral technique~\cite{leordeanu2005spectral} perform spectral analysis and dynamic evolution on the affinity matrix to determine the inlier cluster, respectively. Geometric consistency (GC)~\cite{johnson1998surface,chen20073d} performs inlier cluster selection more straightforwardly. In particular, GC forms a cluster for each correspondence by ensuring correspondences in the cluster are compatible with the query correspondence; the cluster with the maximum element count is served as the inlier cluster. Different from above iterative methods, 3D Hough voting (3DHV)~\cite{tombari2010object} is a one-shot method, which first transforms correspondences to 3D points in a 3D Hough space and then finds the cluster in Hough space.
\begin{figure*}[t]
	\centering
	\includegraphics[width=\linewidth]{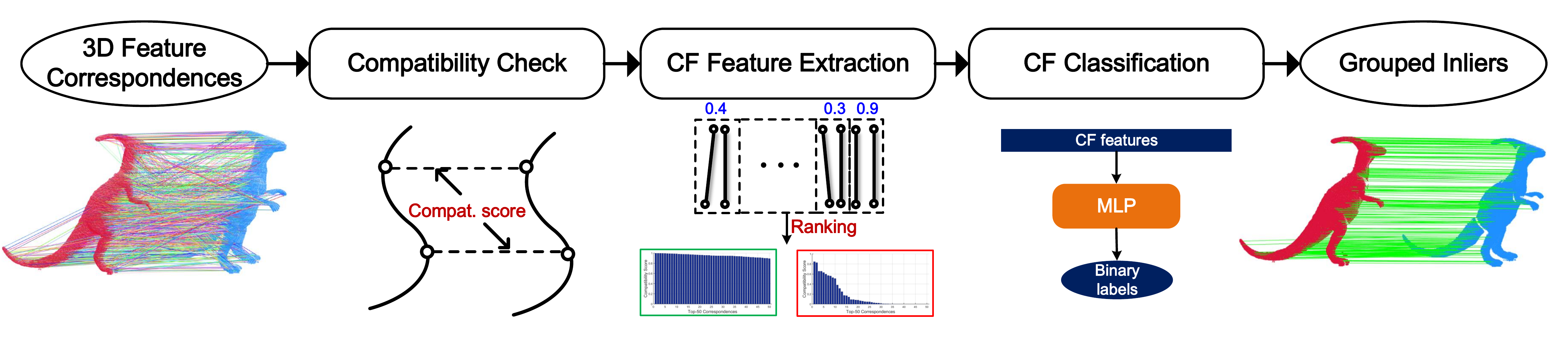}\\
	\caption{Pipeline of the proposed method. {\textit{Compatibility check}}: computing the compatibility scores of a correspondence with others; {\textit{CF feature extraction}}: parameterizing each correspondence by a distinctive CF feature; {\textit{CF classification}}: classifying CF features as inliers and outliers with an MLP network.}
	\label{fig:pipeline}
\end{figure*}

As demonstrated in a recent evaluation~\cite{Yang2020corr_group_eval}, {\textit{group-based methods often miss isolated inliers and are sensitive to low inlier ratios.}}
\\\\\noindent\textbf{Individual-based methods} In early studies, some individual-based methods group correspondences based on feature distances only~\cite{mian2006three,guo2013rotational}, which are straightforward but rely heavily on the performance of descriptors. To achieve more robust grouping, several voting-based methods have been proposed such as search of inliers (SI)~\cite{buch2014search} and consistency voting (CV)~\cite{yang2019ranking}. The common peculiarity of these methods is that one or more voting sets are first defined and then all voters will cast a vote to each correspondence based on some pre-defined rules. 

Compared with group-based methods, individual-based ones assign scores to correspondences independently and thus can more reliably recall isolated inliers. However, existing individual-based methods still exhibit limited precision performance. {\textit{We note that the proposed method is individual-based as well, but is highly selective with outstanding precision performance.}}

\subsection{Learning for Correspondence Grouping}
Existing 3D correspondence grouping methods are still geometric-based ones~\cite{Yang2020corr_group_eval}. In 2D domains, there exist a few mismatch removal methods based on deep learning~\cite{moo2018learning,ma2019lmr,zhao2019nm,sun2020acne}. Yi et al.~\cite{moo2018learning} presented the first attempt to find inliers with an ``end-to-end'' network. To mine local information, Ma et al.~\cite{ma2019lmr} and Zhao et al.~\cite{zhao2019nm} associated spatial and compatibility-specific neighbors to each correspondence for classifier training, respectively. 

Nonetheless, most of existing learning-based image correspondence grouping methods suffer from the following limitations: {\textbf{1)}} the requirement of a large amount of training matching pairs; {\textbf{2)}} the sensitivity to rotations due to the input of coordinate information; {\textbf{3)}} redundant network architectures. By contrast, {\textit{our method properly interprets the roles of geometric and learning techniques, and can effectively overcome these limitations.}}
\section{Methodology}

The pipeline of our method is presented in Fig.~\ref{fig:pipeline}. It consists of three main steps, including compatibility check, CF feature extraction, and CF classification. They play the following roles in the whole pipeline:
\begin{itemize}
	\item {\bf Compatibility check:} one critical difference between inliers and outliers is that inliers are compatible with each other while outliers are usually incompatible with either inliers or outliers. Checking the compatibility between correspondences is the basis of the following steps.
	\item {\bf CF feature extraction:} CF features are extracted based on the compatibility cue to parametrize 3D feature correspondences and distinguish inliers and outliers.
	\item {\bf CF classification:} we train a classifier to classify CF features extracted for correspondences and accomplish the 3D correspondence grouping goal.
\end{itemize}

To improve readability, we introduce the following notations. Let ${\bf P}^s\in {\mathbb{R}^3}$ and ${\bf P}^t\in {\mathbb{R}^3}$ be the source point cloud and the target point cloud, respectively. A feature correspondence set ${\bf C}\in {{\mathbb{R}^6}}$ can be generated by matching local geometric descriptors for ${\bf P}^s$ and ${\bf P}^t$. The aim of our method is to assign a binary label (inlier or outlier) to each element ${\bf c}=({\bf p}^s,{\bf p}^t)$ in ${\bf C}$, where ${\bf p}^s \in {\bf P}^s$ and ${\bf p}^t \in {\bf P}^t$.
\subsection{Compatibility Check}
\begin{figure}[b]
	\centering
	\includegraphics[width=\linewidth]{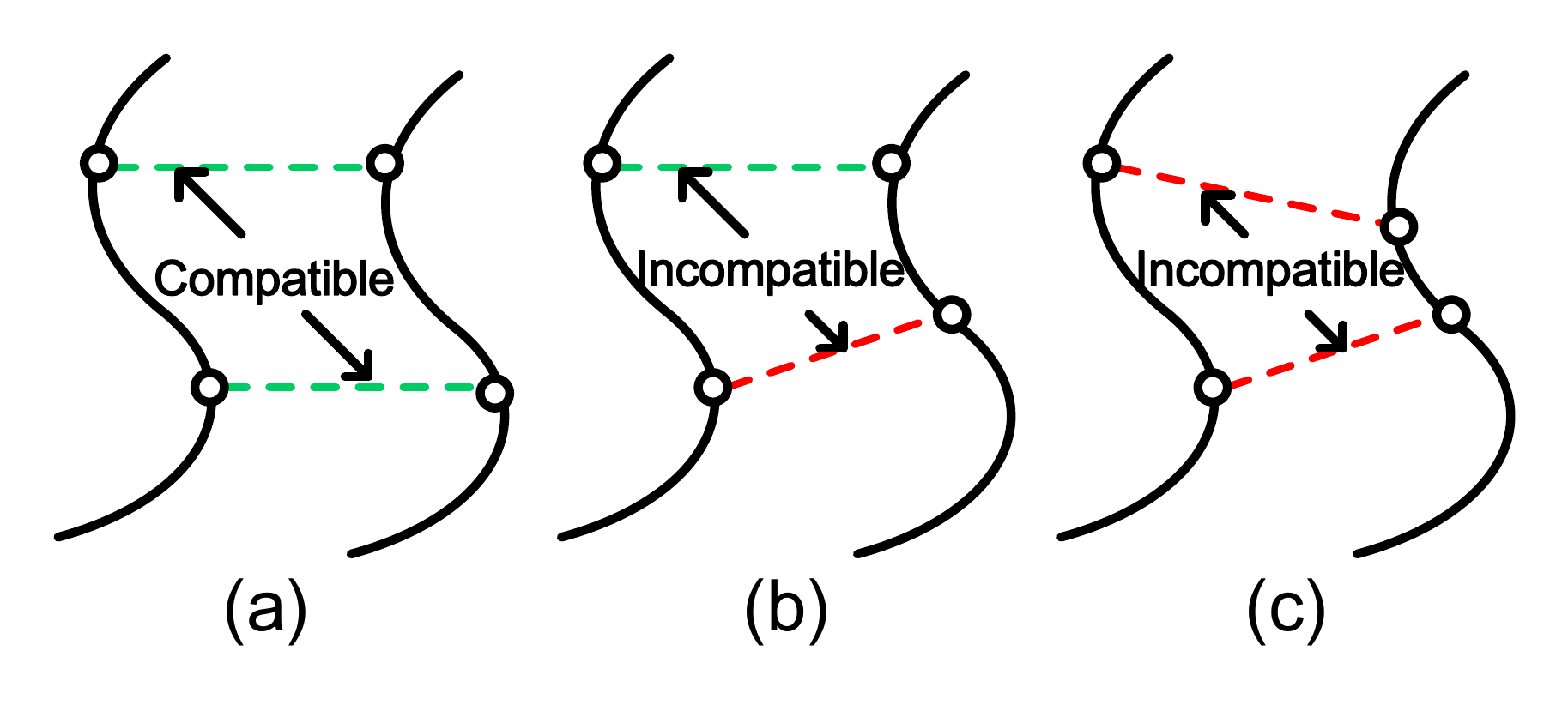}\\
	\caption{Illustration of the statement that (a) inliers are compatible with each other, while ouliers are usually incompatible with either (b) inliers or (c) outliers. Green and red dashed lines denote inliers and outliers, respectively.}
	\label{fig:compt_illus}
\end{figure}
In order to distinguish inliers and outliers, we should fully mine the consistency information within inliers. As depicted in Fig.~\ref{fig:compt_illus}, an important observation is that inliers are geometrically compatible with each other, while outliers are unlikely to be compatible with either outliers or inliers, because the spatial distribution of outliers are unordered. Following this cue, we are motivated to define a metric to check the compatibility between two correspondences.

In the context of 3D point cloud matching, we consider distance and angle constraints~\cite{buch2014search,yang2019ranking} that are invariant to rotations for compatibility metric definition. Let $\bf n$ be the normal of $\bf p$, the distance and angle constraints for two correspondences $({\bf c}_i,{\bf c}_j)$ are respectively defined as:
\begin{equation}
{s}_{dist}({\bf c}_i,{\bf c}_j)=\left|||{\bf p}^s_i-{\bf p}^s_j||-||{\bf p}^t_i-{\bf p}^t_j||\right|,
\end{equation}
and 
\begin{equation}
{s}_{ang}({\bf c}_i,{\bf c}_j)=\left|{\rm acos}({\bf n}^s_i\cdot{\bf n}^s_j)-{\rm acos}({\bf n}^t_i\cdot{\bf n}^t_j) \right|.
\end{equation}
We note that ${s}_{dist}$ and ${s}_{ang}$ are calculated based on linear operation on relative distances and angles, thus being rotation-invariant. Both constraints are complementary to each other (Sect.~\ref{subsec:anay}). By integrating the two constraints, we define the compatibility metric as:
\begin{equation}\label{eq:compt}
S({\bf c}_i,{\bf c}_j)={\rm exp}(-\frac{{s_{dist}}({\bf c}_i,{\bf c}_j)^2}{2\alpha_{dist}^2}-\frac{{s_{ang}}(c_i,c_j)^2}{2\alpha_{ang}^2}),
\end{equation}
where $\alpha_{dist}$ and $\alpha_{ang}$ represent a distance parameter and an angle parameter, respectively. One can see that $S({\bf c}_i,{\bf c}_j) \in [0,1]$ and $S({\bf c}_i,{\bf c}_j)$ equals 1 only if both constraints are fully satisfied. 
\begin{figure}[t]
	\centering
	\includegraphics[width=\linewidth]{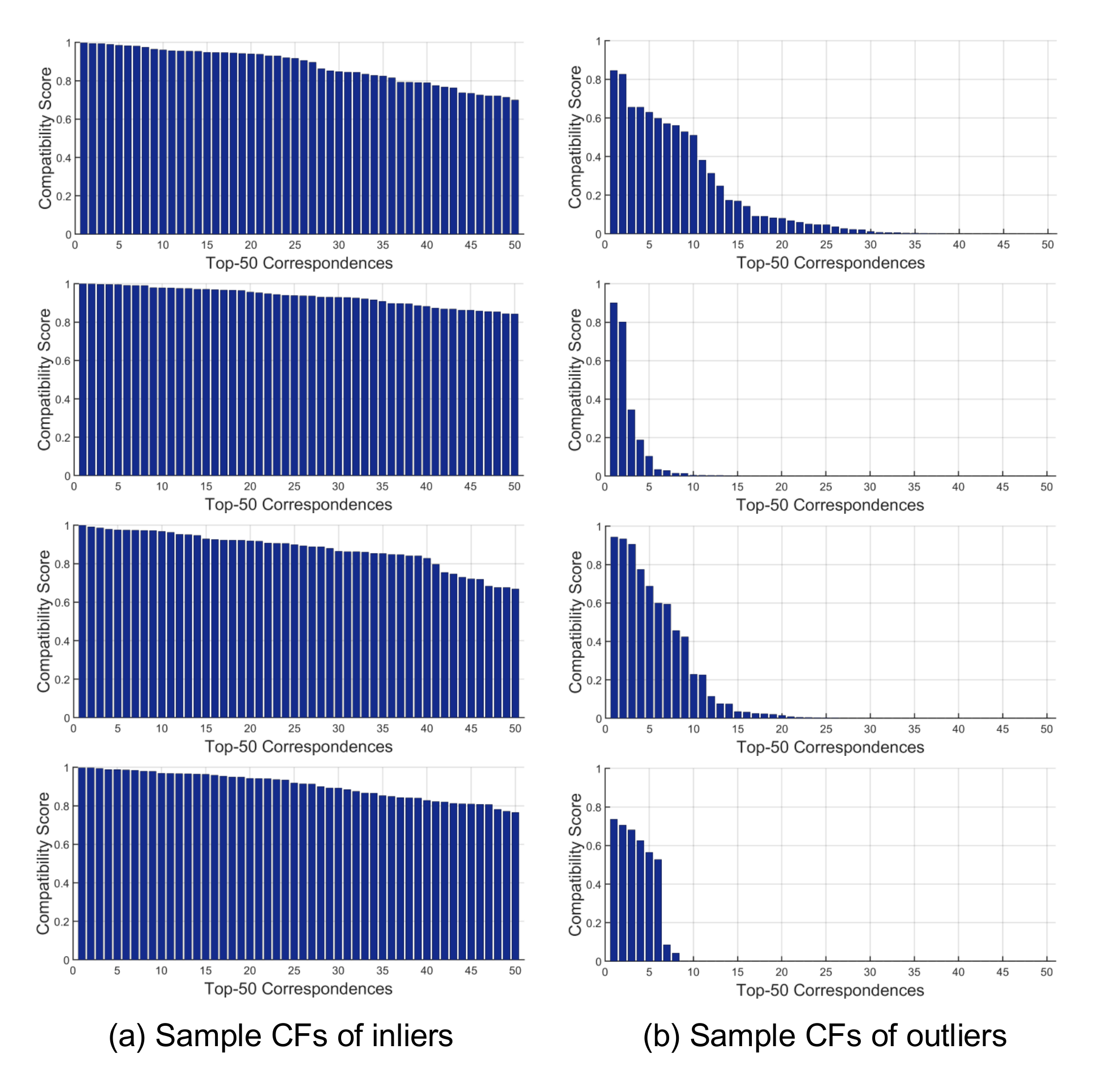}\\
	\caption{Sample CF features of (a) inliers and (b) outliers. We find that with the metric defined in Eq.~\ref{eq:compt} and a proper dimensionality $N$ (50 in the figure), the generated CF features are quite distinctive and intuitively classifiable.}
	\label{fig:compt_features}
\end{figure}
\subsection{CF Feature Extraction}
With a compatibility metric, a naive way for  correspondence grouping is to first assess the greatest compatibility score of each correspondence to others and then set a threshold to filter those with low scores. This is not robust and the distinctiveness of a single compatibility score is limited, as demonstrated in~\cite{chen20073d}. Instead, we consider top-$k$ compatibility scores and render them as a feature vector. Remarkably, most prior works focus on assign scores to correspondences, and the main difference among them is the scoring functions. Our method differs from those ones as we exact feature vectors for correspondences.

Specifically, the calculation of CF features consists of three steps: {\textbf{1)}} compute the compatibility scores  of ${\bf c}$ to other correspondences in $\bf C$ based on Eq.~\ref{eq:compt}, obtaining a score set ${F}=\{S({\bf c},{\bf c}_1),\cdots,S({\bf c},{\bf c}_{D-1})\}$ ($D$ being the cardinality of $\bf C$); {\textbf{2)}} sort elements in $F$ by a descending order, resulting in ${\bf F}=\left[ {\begin{array}{*{10}{c}}{S({\bf c},{\bf c}'_1)}&\cdots&{S({\bf c},{\bf c}'_{D-1})}	\end{array}} \right]$; {\textbf{3)}} compute the $N$-dimensional CF feature ${\bf f}({\bf c})$ of $\bf c$ as the concatenation of the former $N$ elements in $\bf F$, i.e., ${\bf f}({\bf c})=\left[{\begin{array}{*{10}{c}}{{\bf F}(1)}&\cdots&{{\bf F}(N)}	\end{array}} \right]$.

Assume that: {\textbf{1)}} an ideal compatibility scoring metric is defined, which assigns `1' to correspondence pairs composed by inliers and `0' to those with at least one outlier, and {\textbf{2)}} a proper $N$ is defined, we can obtain CF features with all elements being `1' and `0' for inliers and outliers, respectively. Hence,  {\textit{from the theoretic perspective, our proposed CF can be ultra distinctive.}} At present, robust compatibility metric definition for 3D correspondences is still an open issue~\cite{yang2019ranking} and estimating a proper $N$ appears to be a chicken-and-egg problem, resulting in {\textit{noise}} in CF features. However, with the metric defined in Eq.~\ref{eq:compt} and an empirically determined $N$ (based on experiments in Sect. ~\ref{subsec:anay}), {\textit{our CF features, in real case, still hold strong distinctiveness,}} as shown in Fig.~\ref{fig:compt_features}.

\subsection{CF Classification}
\begin{figure}[t]
	\centering
	\includegraphics[width=\linewidth]{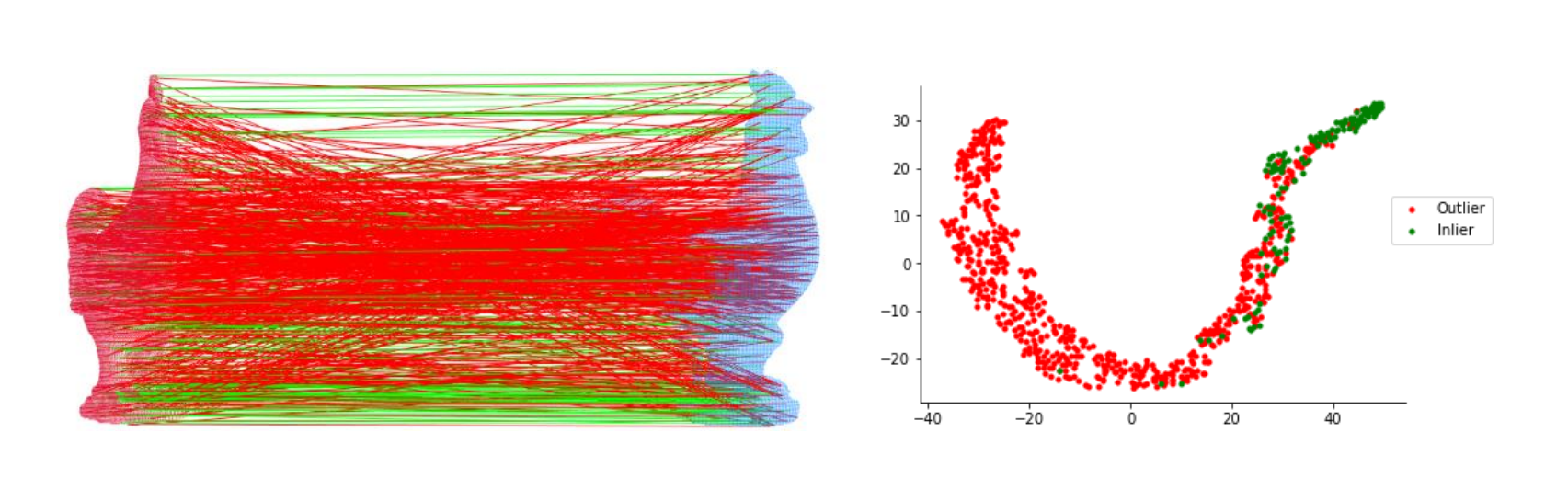}\\
	\caption{Classifying CF features in cases with low inlier ratios appears to be a non-linear classification problem. Left: feature correspondences between two 3D point clouds, where green lines and red lines represent inliers and outliers, respectively. Right: the CF features of all correspondences are projected in a 2D space with t-SNE~\cite{maaten2008visualizing}.}
	\label{fig:sample_tsne}
\end{figure}
Finally, the 3D correspondence grouping problem boils down to a binary feature classification problem. In recent years, deep learning has achieved remarkable success in classification tasks~\cite{deng2009imagenet,krizhevsky2012imagenet}. In addition, we find that classifying CF features in cases with low inlier ratios sometimes appears to be a non-linear classification problem. As shown in Fig.~\ref{fig:sample_tsne}, the CF features of inliers and outliers cannot be linearly separated. Thus, we are motivated to employ a deep-learning classifier.

In particular, the MLP network is suffice to our task because CF feature vectors are inputs to the network. {\textit{This makes the network ultra lightweight}} as compared with other networks for image correspondence problem~\cite{moo2018learning,zhao2019nm,sun2020acne}, which is also demonstrated to be quite effective as will be verified in the experiments. The employed MLP network has 6 layers with 50, 128, 128, 64, 32, and 2 neurons, respectively. Regarding the loss function, we have considered both cross-entropy loss and focal loss~\cite{lin2017focal} (Sect.~\ref{subsec:anay}). We note that the training samples of inliers and outliers are imbalanced for 3D correspondence grouping problem, and eventually we use focal loss to train our network.
\begin{table*}[t]\small
	\centering	
	\scalebox{1}{
		\begin{tabular}{ccccccc}
			\hline
			\bf{Dataset} & \bf{Scenario}& \bf{Nuisances} & \bf{Modality}& \bf{\# Matching Pairs}& \bf Avg. inlier ratio&\\
			\hline
			U3M~\cite{mian2006novel} & Registration & Limited overlap, self-occlusion & LiDAR & 496&0.1480\\
			BMR~\cite{salti2014shot} & Registration & Limited overlap, self-occlusion, real noise & Kinect & 485&0.0563\\
			U3OR~\cite{mian2006three,mian2010repeatability} & Object recognition & Clutter, occlusion & LiDAR& 188&0.0809\\
			BoD5~\cite{salti2014shot}& Object recognition & Clutter, occlusion, real noise, holes & Kinect & 43&0.1575\\
			\hline
		\end{tabular}}
		\caption{Experimental datasets and their properties.}
		\label{tab:dataset}
	\end{table*}
\section{Experiments}
This section presents the experimental setup, analysis and comparative results, along with necessary explanations.
\subsection{Experimental Setup}
\subsubsection{Datasets}
Four datasets are considered in our experiments, including UWA 3D modeling (U3M)~\cite{mian2006novel}, Bologna Mesh Registration (BMR)~\cite{salti2014shot}, UWA 3D object recognition (U3OR)~\cite{mian2006three,mian2010repeatability}, and Bologna Dataset5 (BoD5)~\cite{salti2014shot}. The main properties of experimental datasets are summarized in Table~\ref{tab:dataset}. These datasets have {\bf 1)} different application scenarios, {\bf 2)} a variety of nuisances, and {\bf 3)} different data modalities, which can ensure a comprehensive evaluation. For each dataset, we use correspondence data generated by 75\% matching pairs for training and the remaining for testing. Note that we will also test the generalization performance of our method without training a model for each dataset.
\subsubsection{Metrics}
Precision (P), Recall (R), and F-score (F) are popular metrics for evaluating the performance of correspondence grouping~\cite{zhao2019nm,yang2017performance,Yang2020corr_group_eval}. A correspondence ${\bf c}=({\bf p}^s,{\bf p}^t)$ is judged as correct if:
\begin{equation}
||{\bf p}^s{\bf R}_{gt}+{\bf t}_{gt}-{\bf p}^t||< d_{inlier},
\end{equation}
where $d_{inlier}$ is a distance threshold;  ${\bf R}_{gt}$ and ${\bf t}_{gt}$ denote the ground-truth rotation matrix and translation vector, respectively. We set $d_{inlier}$ to 5 pr as in~\cite{yang2017performance,Yang2020corr_group_eval}. The unit `pr' denotes the point cloud resolution, i.e., the average shortest distance among neighboring points in the point cloud. Thus, precision is defined as:
\begin{equation}
{\rm P}=\frac{|{\bf C}_{inlier}|}{|{\bf C}_{group}|},
\end{equation}
and recall is defined as:
\begin{equation}
{\rm R}=\frac{|{\bf C}_{inlier}|}{|{\bf C}_{inlier}^{gt}|},
\end{equation}
where ${\bf C}_{group}$, ${\bf C}_{inlier}$, and ${\bf C}_{inlier}^{gt}$ represent the grouped inlier set by a grouping method, the true inlier subset in the grouped inlier set, and the true inlier subset in the raw correspondence set. F-score is given by ${\rm F}=\frac{{\rm P}{\rm R}}{{\rm P}+{\rm R}}$. 

We note that 3D correspondence grouping methods are typically applied to rigid registration tasks, e.g., point cloud registration and 3D object recognition, which require sparse and accurate correspondences~\cite{guo20143d}. {\textit{Thus, the precision performance is more critical to these practical applications.}}
\subsubsection{Implementation Details}
For our method, the compatibility check and CF feature exaction modules are implemented in the point cloud library (PCL)~\cite{rusu20113d}, and the MLP classifier is trained in PyTorch with a GTX1050 GPU. The network is optimized via stochastic gradient descent (SGD) with a learning rate of 0.02. All evaluated methods in~\cite{Yang2020corr_group_eval} are compared in our experiments, including similarity score (SS), nearest neighbor similarity ratio (NNSR)~\cite{lowe2004distinctive}, spectral technique (ST)~\cite{leordeanu2005spectral}, random sampling consensus (RANSAC)~\cite{fischler1981random}, geometric consistency (GC)~\cite{chen20073d}, 3D Hough voting (3DHV)~\cite{tombari2010object}, game theory matching (GTM)~\cite{rodola2013scale}, search of inliers (SI)~\cite{buch2014search}, and consistency voting (CV)~\cite{yang2019ranking}. 

To generate 3D feature correspondences between point clouds, we employ the Harris 3D (H3D) detector~\cite{sipiran2011harris} for keypoints detection and the signatures of histograms of orientations (SHOT)~\cite{tombari2010unique} descriptor for local geometric feature extraction. By matching SHOT descriptors via $L_2$ distance, we can obtain initial correspondences. It has been verified in~\cite{Yang2020corr_group_eval} that H3D+SHOT can generate correspondences with {\textit{different spatial distributions, different scales, and different inlier ratios}}, enabling a thorough evaluation.
\subsection{Method Analysis}\label{subsec:anay}
The following experiments were conducted on the U3M dataset (the largest scale one) to analyze the rationality, peculiarities, and parameters of our method.
\\\\\noindent\textbf{Dimensionality of CF features} The dimensionality $N$ of CF features is a key parameter of the proposed method. We test the performance of our method with $N$ being 10, 20, 50, 100, and 200, respectively. The results are shown in Table~\ref{tab:dim}. 
\begin{table}[t]\small
	\renewcommand{\arraystretch}{1}	
	\centering
	\begin{tabular}{c|ccccc}
		\hline
		&\bf 10&\bf 20&\bf 50&\bf 100&\bf 200\\
		\hline
		P  &0.8031&0.7625&0.7483&0.7386&0.7468\\
		
		R  &0.4754&0.5364&0.5308&0.5114&0.4870\\
		
		F  &0.5973&0.6298&0.6211&0.6044&0.5896\\
		\# Epochs  &77&44&7&15&9\\
		\hline
	\end{tabular} 
	\caption{Performance of our method when varying the dimensionality of CF features.}
	\label{tab:dim}
\end{table} 
\begin{table*}[t]\small
	\renewcommand{\arraystretch}{1}
	\centering
	\begin{tabular}{c|cccccccccc}
		\hline
		&\bf CE(1:1)& \bf CE(1:4)&\bf CE(1:8)&\bf CE(1:10)&\bf CE(raw)&\bf FL(1:1)&\bf FL(1:4)&\bf FL(1:8)&\bf FL(1:10)&\bf FL(raw)\\
		\hline
		P  &0.2893&0.4149&0.5688&0.6120&NC&0.2431&0.4362&0.5510&0.6180&0.7483\\
		
		R  &0.8615&0.7828&0.6736&0.6439&NC&0.8827&0.7692&0.6877&0.6394&0.5308\\
		
		F  &0.4332&0.5424&0.6168&0.6275&NC&0.3812&0.5567&0.6118&0.6285&0.6210\\
		\hline
	\end{tabular}
	\caption{Comparison of cross entropy loss (CE) and focal loss (FL) when varying the ratio of positive sample count to negative sample count (NC: not converge; raw: the ratio is about 1:25 in raw training data).}
	\label{tab:loss}
\end{table*}
The results indicate that $N=20$ and $N=50$ achieve the best and the second best performance, respectively. Thus, a proper $N$ is needed to maximize the distinctiveness between the CF features of inliers and outliers. In addition, we find that the network converges much faster with $N=50$ than other settings, and we set $N$ to 50 by default.  
\\\\\noindent\textbf{Focal loss vs. cross entropy}
To prepare training data, we have two alternatives: using {\textit{equal}} or {\textit{imbalanced} numbers of positive samples and negative samples. The later one is closer to real matching case. Here, we compare the cross entropy loss and focal loss when varying the ratio of positive sample count to negative sample count. The results are reported in Table~\ref{tab:loss}.

One can see that the performance of both losses improves when ratio of positive samples to negative samples decreases from 1:1 to 1:10, and their gap is marginal. When more negative samples are included (i.e., all samples in raw training data), focal loss achieves better precision performance while the network with cross entropy loss fails to converge. As expected, focal loss is more suitable to 3D correspondence grouping problem where a large portion of training data are outliers.
\\\\\noindent\textbf{Varying compatibility metrics} A critical factor to the proposed CF features is the definition of compatibility metrics. In our defined compatibility metric (Eq.~\ref{eq:compt}), both distance and angle constraints are considered. Here, we test the effect when using looser constraints, i.e., solely using either distance constraint or angle constraint, as shown in Table~\ref{tab:compt}.
\begin{table}[t]\small
	\renewcommand{\arraystretch}{1}
	
	\centering
	\begin{tabular}{c|ccc}
		\hline
		&\bf Distance&\bf Angle&\bf Both\\
		\hline
		P  &0.6443&NC&0.7483\\
		
		R  &0.6885&NC&0.5308\\
		
		F  &0.6657&NC&0.6211\\
	\hline
	\end{tabular} 
	\caption{The effect of using compatibility metrics with different geometric constraints (NC: not converge).}
	\label{tab:compt}
\end{table} 

It is interesting to see that using a slightly looser constraint (distance only) can achieve better F-score performance than using both constraints. However, when the constraint is too loose (angle only), the network cannot converge because the generated CF features are ambiguous. Because using both constraints achieves the best precision performance, which is preferred in most application scenarios, so we consider both constraints to define the compatibility metric.
\\\\\noindent\textbf{PointNet vs. MLP} As similar to some 2D correspondence methods~\cite{moo2018learning,sun2020acne}, directly setting the coordinates of correspondences as the input to networks can be another way for grouping. We tested the performance of using coordinate information for learning with PointNet~\cite{qi2017pointnet} on testing data with and without arbitrary rotations. The results are reported in Table~\ref{tab:rot}.
\begin{table}[t]\small
	\renewcommand{\arraystretch}{1}
	
	\centering
	\begin{tabular}{c|cccc}
		\hline
		&\bf PointNet&\bf PointNet ($SO(3)$)&\bf Ours&\bf Ours ($SO(3)$)\\
		\hline
		P  &0.3888&0.1290&0.7483&0.7483\\
		
		R  &0.0355&0.0018&0.5308&0.5308\\
		
		F  &0.0651&0.0035&0.6211&0.6211\\
		\hline
	\end{tabular} 
	\caption{Comparison of PointNet~\cite{qi2017pointnet} with coordinates being input and our method with CF features being input on testing data with and without arbitrary $SO(3)$ rotations.}
	\label{tab:rot}
\end{table} 
\begin{table}[t]\small
	\renewcommand{\arraystretch}{1}
	\centering
	\begin{tabular}{c|cccc}
		\hline
		&\bf $\bf \frac{1}{8} \times$ 490k&\bf $\bf \frac{1}{4} \times$ 490k&\bf $\bf \frac{1}{2} \times$ 490k&\bf  490k\\
		\hline
		P  &0.7653&0.7533&0.7558&0.7483\\
		
		R  &0.5130&0.5219&0.5199&0.5308\\
		
		F  &0.6142&0.6166&0.6160&0.6211\\
		\# Epochs  &156&96&48&15\\
		\hline
	\end{tabular} 
	\caption{The effect of varying the amount of training data on our method.}
	\label{tab:num_data}
\end{table} 
\begin{figure*}[t]
	\centering
	\includegraphics[width=0.9\linewidth]{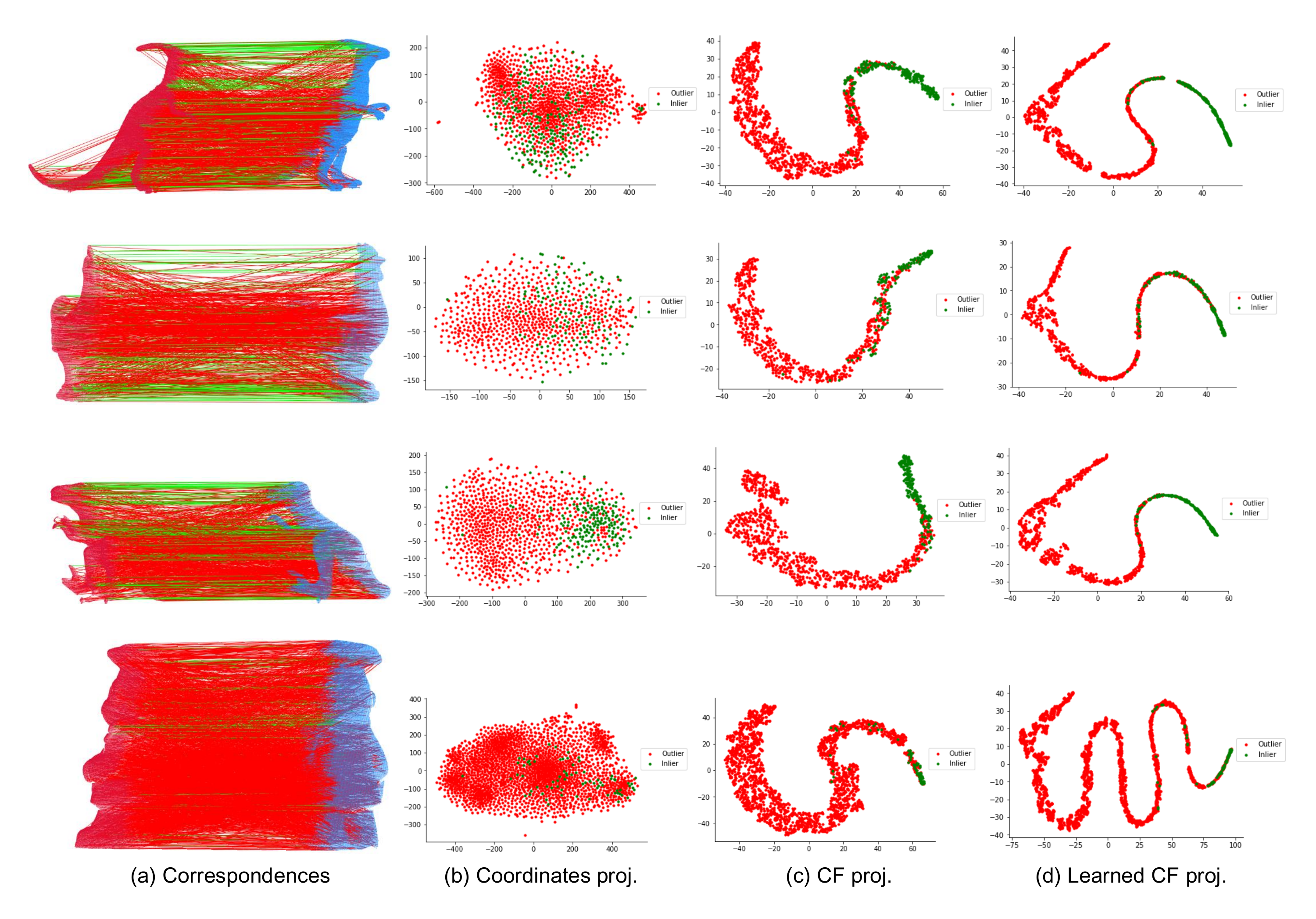}\\
	\caption{Sample results of (a) 3D feature correspondences, and  2D projections (by t-SNE~\cite{maaten2008visualizing}) of (b) correspondence coordinates, (c) CF features, and (d) the features of the second last layer of MLP.}
	\label{fig:coor_feat_tsne}
\end{figure*}
\begin{table*}[t]\small
	\renewcommand{\arraystretch}{1}
	
	\centering
	\begin{tabular}{c|cccccccccc}
		\hline
		&\bf SS&\bf NNSR~\cite{lowe2004distinctive} &\bf ST~\cite{leordeanu2005spectral}&\bf RANSAC~\cite{fischler1981random}&\bf GC~\cite{chen20073d}&\bf 3DHV~\cite{tombari2010object}&\bf GTM~\cite{rodola2013scale}&\bf SI~\cite{buch2014search}&\bf CV~\cite{yang2019ranking}&\bf CF (Ours)\\
		\hline
		\multicolumn{11}{c}{(a) \textit{U3M dataset} } \\
		\hline
		\rowcolor{green!30}P&0.0374&0.1289&{0.3984}&\underline{0.5442}&0.2920&0.1960&0.5285&0.0380&0.1092&\bf 0.7483\\
		R&0.3819&0.4084&0.5833&0.8493&0.7499&0.6999&0.5987&\bf 0.9996&0.9839&0.5308\\
		F&0.0681&0.1960&0.4734&\bf 0.6634&0.4203&0.3062&0.5614&0.0733&0.1966&\underline{0.6211} \\
		\hline
		\multicolumn{11}{c}{(b) \textit{BMR dataset}} \\
		\hline
		\rowcolor{green!30}P&0.0243&0.0606&0.2993&0.3737&0.1458&0.1492&\underline{0.3946}&0.0350&0.0700&\bf 0.8575\\
		R&0.3405&0.0967&0.3734&0.8178&\underline{0.5740}&0.5049&0.3626&0.5522&\bf 0.9438&0.1529\\
		F&0.0454&0.0745&0.3323&\bf 0.5129&0.2325&0.2304&\underline{0.3779}&0.0658&0.1303&0.2596\\
		\hline
		\multicolumn{11}{c}{(c) \textit{BoD5 dataset}} \\
		\hline
		\rowcolor{green!30}P&0.0474&0.1635&0.5660&\underline{0.5961}&0.5207&0.3927&\bf 0.7022&0.0748&0.3593&0.5699\\
		R&0.2024&0.1136&0.4086&0.8747&0.7559&\underline{0.8890}&0.4556&0.7337&\bf 0.9869&0.4151\\
		F&0.0768&0.1341&0.4746&\bf 0.7090&\underline{0.6166}&0.5448&0.5527&0.1359&0.5268&0.4804\\
		\hline
		\multicolumn{11}{c}{(d) \textit{U3OR dataset}} \\
		\hline
		\rowcolor{green!30} P&0.0171&0.0724&0.1119&\underline{0.5812}&0.1918&0.1190&0.4907&0.0143&0.0523&\bf 0.8641\\
		R&0.4111&0.5296&0.1670&0.2442&0.6302&0.3537&0.5224&\bf 1.0000&\underline{0.9461}&0.3196\\
		F&0.0328&0.1274&0.1340&0.3438&0.2941&0.1781&\bf 0.5061&0.0282&0.0991&\underline{0.4666} \\
		\hline
	\end{tabular} 
	\caption{Comparison of the proposed method with nine state-of-the-art methods in terms of precision, recall, and F-score performance on four experimental datasets (bold: the best; underlined: the second best).}
	\label{tab:compare}
\end{table*} 
Two observations can be made from the table. {\textbf{1)}} PointNet with coordinates being the input achieves significantly worse performance than our MLP architecture with CF features being input. This is because the range of 3D real-world coordinate information is too large, which makes the network very difficult to mine the patterns within dataset. {\textbf{2)} Coordinates are sensitive to rotations, making the performance of PointNet even worse when the testing data undergoing rotations. By contrast, because our CF features consist of compatibility scores computed based on rotation-invariant constraints, making CF and the CF-based learning network rotation-invariant as well.
	
To further support our statement, we visualize some exemplar  results of feature correspondences, projections of correspondence coordinates, CF features, and the features of the second last layer of MLP in Fig.~\ref{fig:coor_feat_tsne}. Obviously, one can hardly mine consistencies within inliers from the coordinate information. By contrast, CF features hold strong distinctiveness. In addition, learned CF features by MLP can further enhance the distinctiveness (the clusters of inliers and outliers in Fig.~\ref{fig:coor_feat_tsne}(d) are tighter than these in Fig.~\ref{fig:coor_feat_tsne}(c)).
\\\\\noindent\textbf{Varying the amount of training data}
The initial number of correspondences used for training in the U3M dataset is around 490k. We test the cases with less training data and report the results in Table~\ref{tab:num_data}.

The table suggests that our method behaves well even removing $\frac{7}{8}$ training data, while requiring much more training epochs. We note that dozens of point cloud pairs can generate correspondences at $\frac{1}{8}\times$490k level. As compared with methods relying over tens thousand of matching pairs~\cite{moo2018learning,zhao2019nm}, {\textit{our method can be trained with significantly less matching pairs.}}
\subsection{Comparative Results \& Visualization}
\noindent\textbf{Start-of-the-art comparison} All evaluated methods in a recent evaluation~\cite{Yang2020corr_group_eval} are compared with the proposed method on four experimental datasets. All methods are tested on the same testing data. The results are shown in Table~\ref{tab:compare}.

\begin{table}[t]\small
	\renewcommand{\arraystretch}{1}
	\centering
	\begin{tabular}{c|cccc}
		\hline
		&\bf BMR&\bf U3M+noise&\bf U3M+simplification&\bf ISS+FPFH\\
		\hline
		P  &0.6928&0.7407&0.7088&0.7409\\
		
		R  &0.3241&0.4111&0.3247&0.4342\\
		
		F  &0.4416&0.5287&0.4454&0.5475\\
		\hline
	\end{tabular} 
	\caption{Generalization performance of the proposed method (the model is trained on the original U3M dataset).}
	\label{tab:gene}
\end{table} 
The following observations can be made from the table. {\textbf{1)}} Our method achieves the best precision performance on the U3M, BMR, and U3OR dataset. Moreover, the gap between our method and the second best one is significant on the BMR and U3OR datasets. On the BoD5 dataset, our method is surpassed by GTM and RANSAC. However, this dataset is less challenging than the other three ones (Table~\ref{tab:dataset}).  This indicates that our method can achieve superior precision performance especially on data with low inlier ratios. We also note that only 33 pairs of data are leveraged to train our network on the BoD5 dataset. {\textbf{2)}} In terms of the recall performance, SI and CV, as two typical individual-based methods, achieve top-ranked performance. Unfortunately, their precision performance is quite limited. This could result in inaccurate and time-consuming rigid registration results due to heavy outliers in the grouped inlier set. {\textit{We note that a looser geometric constraint can be used if a balance is needed between precision and recall (as verified in Table~\ref{tab:compt}), indicating that our method is flexible.}} {\textbf{3)}} Although the proposed method is an individual-based one, it is quite selective with superior precision performance. Notably, GTM appears to be the most selective method as evaluated by~\cite{Yang2020corr_group_eval}, while our method generally outperforms it by a large margin in terms of precision.
%The overall superiority of our method can be explained from at least three aspects. First, CF consists of compatibility scores of a candidate to some other correspondences. The compatibility between 3D correspondences is a useful cue for correspondence grouping because inliers are compatible with each other. Thus CF is supposed to be distinctive. Second, although classifying CF features in some cases with extremely low inlier ratios appears to be a nonlinear problem, deep learning techniques can well handle this issue. Moreover, our method only uses a simple MLP network and achieves decent performance. Therefore, {\textit{the advantages of geometric methods and deep learning approaches are properly leveraged in our method.}}
\\\\\noindent\textbf{Generalization performance}
We use the model trained on the initial U3M dataset to predict inliers on the following datasets: BMR dataset, variants of U3M datasets  with 0.3 pr Gaussian noise,  $\frac{1}{8}$ random data decimation, and ``ISS detector~\cite{zhong2009intrinsic} + FPFH descriptor~\cite{rusu2009fast}'', respectively. The results are shown in Table~\ref{tab:gene}. {\textit{One can see that the model trained on the U3M dataset also achieves decent performance when changing the testing dataset, injecting additional nuisances, and changing ``detector-descriptor'' combinations.}} This is potentially because the eventual results caused by above test conditions is the variation in inlier ratios, while our CF features can effectively mine the hidden consistencies of inliers and inconsistencies of outliers in different inlier ratio cases. 
\\\\\noindent\textbf{Visualization}
\begin{figure}[t]
	\centering
	\includegraphics[width=\linewidth]{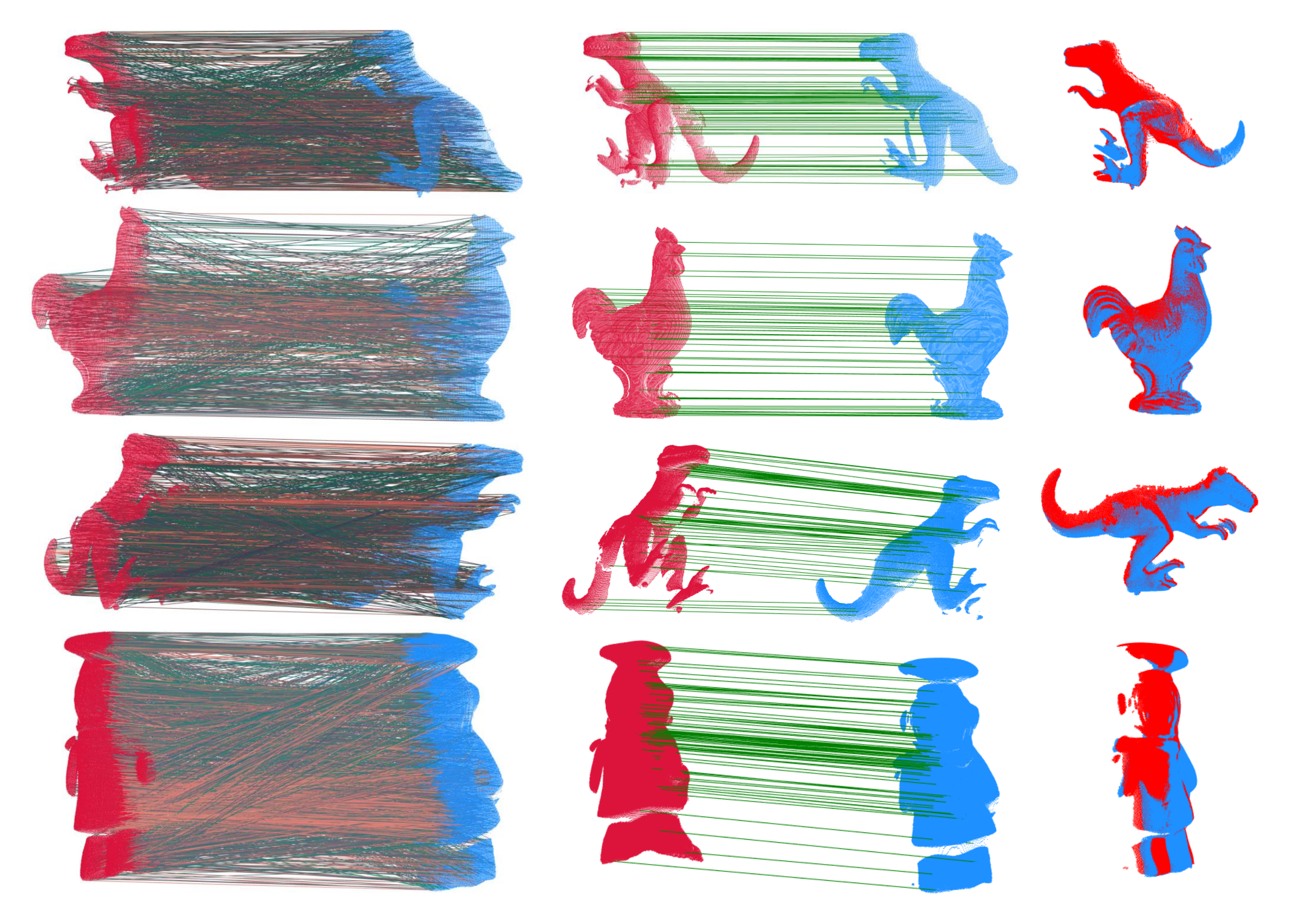}\\
	\caption{Sample visualization results. From left to right: initial correspondences with colors obtained by projecting CF features to 3D RGB space, grouped correspondences by our method, and the registration result with the grouped correspondences using PCL~\cite{rusu20113d}.}
	\label{fig:vis}
\end{figure}
Finally, we give some visualization results of our method in Fig.~\ref{fig:vis}. Two observations can be made. First, the colors of correspondences obtained by projecting CF features to 3D RGB space can reflect the consistency of inliers. Second, the grouped correspondences by our method are quite consistent and can achieve accurate 3D registration results.

\section{Conclusion}
We presented a novel representation to describe 3D feature correspondence named compatibility feature (CF), along with a CF-based 3D correspondence grouping method for 3D correspondence grouping. CF captures the main distinctiveness between inliers and outliers regarding pairwise geometrical compatibility, which is rotation-invariant as well. With CF features, a lightweight MLP network is able to classify them and achieve outstanding performance. Experiments on four standard datasets with a rich variety of  application scenarios and nuisances paired with comparisons with nine state-of-the-art methods demonstrate the overall superiority of our method. We also find that the pipeline of our proposed CF-based 3D correspondence grouping method can be generalized to matching problems for many other data representations, such as 2D images and non-rigid point clouds/meshes, which remains an interesting future research direction.
{\small
	\bibliographystyle{ieee_fullname}
	\bibliography{mybibfile}
}
\end{document}